\documentclass[a4paper]{article}

\title{Key Principles in Cross-Domain Hyper-Heuristic Performance}
\author{Václav Sobotka$^1$, Lucas Kletzander$^2$, Nysret Musliu$^2$, and Hana Rudová$^1$}

\usepackage{amsmath}
\usepackage{amsfonts}
\usepackage{tabularx}
\usepackage{booktabs}
\usepackage{makecell}
\usepackage{bm}
\usepackage{multirow}
\usepackage{caption}
\usepackage{times}
\usepackage{helvet}
\usepackage{courier}
\usepackage{graphicx}
\usepackage{pdflscape}

\usepackage{natbib}
\date{
    sobotka@mail.muni.cz, lucas.kletzander@tuwien.ac.at, nysret.musliu@tuwien.ac.at, hanka@fi.muni.cz \\[1em]
    $^1$ Faculty of Informatics, Masaryk University, Brno, Czech Republic \\
    $^2$ Christian Doppler Laboratory for Artificial Intelligence and Optimization for Planning and Scheduling, DBAI, TU Wien, Austria
}      

\begin{document}

\maketitle

\begin{abstract}
    Cross-domain selection hyper-heuristics aim to distill decades of research on problem-specific heuristic search algorithms into adaptable general-purpose search strategies. In this respect, existing selection hyper-heuristics primarily focus on an adaptive selection of low-level heuristics (LLHs) from a \textit{predefined} set. In contrast, we concentrate on the composition of this set and its strategic transformations. We systematically analyze transformations based on three key principles: solution acceptance, LLH repetitions, and perturbation intensity, i.e., the proportion of a solution affected by a perturbative LLH. We demonstrate the raw effects of our transformations on a trivial unbiased random selection mechanism. With an appropriately constructed transformation, this trivial method outperforms all available state-of-the-art hyper-heuristics on three challenging real-world domains and finds 11 new best-known solutions. The same method is competitive with the winner of the CHeSC competition, commonly used as the standard cross-domain benchmark. Moreover, we accompany several recent hyper-heuristics with such strategic transformations. Using this approach, we outperform the current state-of-the-art methods on both the CHeSC benchmark and real-world domains while often simplifying their designs.
\end{abstract}

\section{Introduction}

Metaheuristic search strategies~\cite{METAHEURISTICS_handbook} are the cornerstone of methodologies tackling a vast range of combinatorial optimization problems. While metaheuristics address issues common to all search algorithms, such as local optima evasion or balancing exploitation and exploration in the process, they still serve as templates for implementations of problem-specific algorithms. In contrast, selection hyper-heuristics~\cite{SURVEY_hyperheuristics_2020} aim to provide domain-agnostic search strategies. Given a set of low-level heuristics (LLHs), hyper-heuristics aim to steer the search process by adaptively selecting from the available LLHs, often using information from previous iterations of the search process. Research on hyper-heuristics has attracted attention, arguably due to the appealing idea of generalizing and consolidating decades of research on problem-specific approaches and metaheuristics into general domain-agnostic methods. As a result, a wide range of adaptive and learning LLH selection mechanisms has emerged.

Interestingly, the vast majority of works on hyper-heuristic methods assume a fixed set of LLHs is provided and use it as is. In this paper, we challenge this approach by transparently transforming existing LLH sets into new \textit{virtual LLH sets}. Our virtual LLH sets effectively modify the original LLHs in three fundamental aspects: solution acceptance, repeated LLH applications, and perturbation intensity. By running existing hyper-heuristics on top of properly designed virtual LLH sets, we systematically obtain significant cross-domain performance benefits. With this approach, we demonstrate substantial improvements for the majority of the available recent hyper-heuristics, outperforming the current state-of-the-art hyper-heuristics on both the standard CHeSC cross-domain benchmark~\cite{CHeSC_competition} and three challenging real-world application domains. Critically, we exemplify the raw effects of the aforementioned three key principles on a trivial random unbiased selection mechanism. We show that the trivial selection mechanism, accompanied by a strategically transformed LLH set, outperforms all tested state-of-the-art hyper-heuristics on three real-world domains and provides performance comparable to the CHeSC-winning hyper-heuristic~\cite{CHeSC_GIHH} on the standard CHeSC benchmark. To summarize, the key contributions of our paper are:
\begin{itemize}
    \item We identify and analyze the three aforementioned key principles, use them to transparently transform existing LLH sets, and demonstrate their critical impacts on cross-domain search performance.
    \item We demonstrate that solely the three key principles accompanying a trivial selection mechanism are enough to obtain results comparable to or even outperforming recent hyper-heuristics. This finding strongly contrasts with the general focus on the LLH selection mechanisms.
    \item Using only the trivial selection mechanism and the three key principles, we obtain 11 new best-known solutions on two challenging real-world domains.    
    \item By transforming existing LLH sets, we significantly improve the cross-domain performance for the majority of the available recent hyper-heuristics. We demonstrate the benefits of our methodology on both standard benchmarks and three real-world domains.
\end{itemize}

\section{Related works}\label{section:related_works}

Hyper-heuristics aim to provide high-level search strategies. Previously, comprehensive reviews and classifications of existing hyper-heuristics were provided in \cite{SURVEY_hyperheuristics_2013,SURVEY_hyperheuristics_2020}, and most recently by \cite{SURVEY_hyperheuristics_2024}. In terms of the standard classification \cite{burke2019classification}, we concentrate on \textit{perturbative selection hyper-heuristics}. Such hyper-heuristics operate \textit{online} by progressively selecting the LLHs to be applied and adapting this selection based on the past search trajectory. A large body of research concentrating on this type of hyper-heuristics is centered around the Cross-Domain Heuristic Search Challenge 2011 (CHeSC) \cite{CHeSC_competition}. The competition introduced the HyFlex framework \cite{HYFLEX_benchmark}, providing a diverse benchmark implementing six search domains with a standardized interface and predefined LLH sets for each domain. The CHeSC competition-winning hyper-heuristic GIHH \cite{CHeSC_GIHH} combines a large number of adaptive mechanisms. It was followed by a self-adaptive variable-neighborhood search hyper-heuristic \cite{CHeSC_VNS}, and a method using intensification-diversification cycles with reinforcement learning mechanisms \cite{CHeSC_ML}. Overall, there was a total of 20 teams competing in CHeSC, resulting in a collection of well-evaluated hyper-heuristic concepts and mechanisms. Notably, a recent meta-study \cite{METASTUDY_2025} systematically reviews all hyper-heuristics competing in CHeSC, comprehensively analyzing several key design decisions and their impact on the methods' performance. 

Among the hyper-heuristics developed after CHeSC, we observe several trends. First is the aim for design simplicity. In~\cite{2016_LGIHH} (LGIHH), the original GIHH algorithm was simplified by analyzing its mechanisms and eliminating the unnecessary ones. The simplified algorithm was shown to outperform the original GIHH algorithm. With a similar motivation, \cite{2014_FSILS} (FSILS) introduced a conceptually simple algorithm combining an iterated local search (ILS) scheme with time normalization, acceptance, and restart mechanisms. FSILS is shown to outperform GIHH with a strikingly simpler design and a well-documented role of all applied mechanisms. Subsequently, the key design decisions of FSILS were later used in \cite{2021_TSILS} (TSILS) with adaptive mechanisms combined with Thompson sampling LLH selection, and in \cite{2022_EAILS} (EAILS), where evolutionary mechanisms steer the combination of perturbative LLHs. To the best of our knowledge, the results of TSILS on the CHesC benchmark form the current state-of-the-art. The last notable trend is the employment of learning, often based on reinforcement learning techniques~\cite{RL_sutton_barto} (RL). In \cite{2014_GEPHH} (GEPHH), gene expression programming is used to automatically select LLHs and acceptance mechanisms. The Monte Carlo tree search scheme with multi-armed bandit principles was used in \cite{2015_MCTS} (MCTS) and \cite{2017_FRAMAB_RL} (FRAMAB) to steer the search trajectory. Later, both \cite{2018_QHH} (QHH) and \cite{2022_MC_RL} (MC) used Q-learning-based selection of LLHs. Most recently, \cite{2023_LARGE_STATE_RL} (LASTRL) combined ILS with adaptive RL strategies and rich state representation.

We separately emphasize~\cite{chuang_thesis} (LUBY) using Luby sequence restarts~\cite{LUBY_sequence} in an automated synthesis of search strategies. To the best of our knowledge, this is the only existing work proposing an LLH set transformation. The described \textit{domain amplification} doubles the LLH set size by adding an "amplified" duplicate for each LLH. The amplified LLHs execute the original LLH for 10 ms while rejecting non-improving solutions. Interestingly, this exact concept was later used in \cite{2022_MC_RL} with substantial performance benefits. Still, the only available description of this technique known to us \cite[p.~40-41]{chuang_thesis} provides minimal justification of its design. The work motivates the repeated LLH applications with the aim to "promote collaboration among heuristics". Yet, if worse solutions are always discarded, such collaboration and the ability to escape local optima are vastly limited. Therefore, we cover the described gap by systematically examining LLH set transformations that introduce solution acceptance and repeated LLH applications (as in domain amplifications), and further add a third important principle, perturbation intensity. Table~\ref{table:hh_properties} concludes our review by summarizing all post-CHeSC methods focusing on these three principles and design features strongly influencing methods' exploration-exploitation balance. We later compare against all methods in Table~\ref{table:hh_properties}.

\begin{landscape}
\begin{table*}[tb]
        \small
        \centering
        \begin{tabular}{>{\small}c >{\small}c >{\small}c >{\small}c >{\small}c >{\small}c }
        \makecell[bc]{\textit{Work}} & \textit{Acceptance mechanism} & \textit{LLH chaining} & \textit{Perturbation intensity} & \textit{LLH selection bias} & \textit{Restarts} \\
        \toprule
        \textbf{LASTRL} & All & LS-biased chains & Static & RL, ILS & Luby, Full \\
        \textbf{MC} & All, Discard worse & Repeat until timeout & Static & RL & Luby \\
        \textbf{EAILS} & $\mu$-norm Metropolis & LS chains & Adaptive & EA-learned ILS & -- \\
        \textbf{TSILS} & $\mu$-norm Metropolis & LS chains & Adaptive & Double shaking ILS & -- \\
        \textbf{LUBY} & All, Discard worse & Repeat until timeout & Static & Random & Luby \\
        \textbf{QHH} & $\mu$-norm Metropolis + others & LS chains & Not stated & RL, ILS & - \\
        \textbf{FRAMAB} & Monte Carlo & -- & Adaptive & Multi-armed bandit & - \\
        \textbf{LGIHH} & AILLA & Relay hybridization & Adaptive & Adaptive LLH selection & Full \\
        \textbf{MCTS} & Monte Carlo & -- & Multiple static & Multi-armed bandit & -- \\
        \textbf{FSILS} & $\mu$-norm Metropolis & LS chains & Static & ILS & Full \\
        \textbf{GEPHH} & Evolving function & -- & Not stated & Evolutionary & -- \\
        \bottomrule
        \end{tabular}
        \vspace*{-1.25ex}
        \caption{
        Key design mechanisms affecting the exploration-exploitation balance in recent cross-domain hyper-heuristics.}
        \label{table:hh_properties}
        \vspace*{-1.75ex}
\end{table*}
\end{landscape}

\section{LLH set transformation framework}~\label{sec:technical}
Our work implements and publishes the proposed LLH set transformation methodology as an extension module of HyFlex~\cite{HYFLEX_benchmark}, a commonly used framework for the development and benchmarking of cross-domain selection hyper-heuristics. Thus, we frame the descriptions of our methodology using interfaces and terms tied to HyFlex. In HyFlex, a \textit{problem domain} must specify a set of LLHs. Furthermore, the domain maintains a number of solutions in addressable \textit{solution registers}. The stored solutions can be copied between the registers and inspected for solution costs. The hyper-heuristics then apply the available LLHs to these solutions and manipulate the solutions in the registers to steer the search trajectory. Further, the hyper-heuristics may instruct the domain object to alter its parameter controlling the perturbation intensity. From the hyper-heuristic point of view, the only available information about each LLH is its unique identification and membership in one of the following four categories. \textit{Local search} (LS) LLHs are typically classical neighborhood-based search moves with the additional guarantee that their application to a solution returns a solution of equal or better quality. \textit{Ruin\&recreate} (RR) LLHs "ruin" the solution first, making it a partial solution. Then, the partial solution is "recreated", typically in a greedy manner. \textit{Mutation} (MUT) LLHs only aim to introduce random changes to the solution. \textit{Crossover} (XO) LLHs combine two existing solutions into a new one. Neither RR, MUT, nor XO LLHs provide solution quality guarantees. Often, the RR and MUT categories are jointly called \textit{perturbative LLHs}. The LLH set and its division into these four categories form the input to our transformation procedures.

The key idea of our approach is to transform an original LLH set into a new virtual LLH set populated with \textit{virtual LLHs} and let existing hyper-heuristics transparently operate on top of it. Our transformations target the three previously discussed principles, and we apply the transformations at the level of individual LLH categories. Within the virtual LLH set, each virtual LLH is based on one LLH from the original set (attribute \textsc{LLH}). The virtual LLH attributes \textsc{Accept}, \textsc{Duration}, and \textsc{Intensity} then describe the modifying effects in terms of the three principles as per the applied transformation. Lastly, we note that it is possible to apply several transformations on one LLH category, as we commonly do to obtain several LLH duplicates with different perturbation intensity.
\begin{figure}[tb]
     \begin{tabbing}
         \ \,\,\,1: \=\textbf{procedure} \textsc{ApplyVirtualLLH}($D, H, s^{*}, s^{**}$)\\
         \> {\it$D$ \qquad \qquad \,\, HyFlex problem domain}\\
         \> {\it$H$  \qquad \qquad \,\, virtual LLH}\\
         \> {\it$s^{*}$ \qquad \qquad \,\, source solution register ID}\\
         \> {\it$s^{**}$ \qquad \quad \,\,\,\,\, target solution register ID}\\
         \> {\it \boldsymbol{$\mu$} \qquad \qquad \,\,\, mean improvement statistic \textbf{(global)}}\\
         \> {\it \boldsymbol{$n_{imp}$} \qquad \,\,\,\, number of improvements statistic \textbf{(global)}}\\[1ex]
         \ \,\,\,2: \> \quad \= $D$.\textsc{CopySolution}($s^{*}$, $s_{\mathrm{cur}}$) \\
         \ \,\,\,3: \> \> $D$.\textsc{SetPerturbationIntensity}($H$.\textsc{Intensity}) \\
         \ \,\,\,4: \> \>\textbf{while} $H$.\textsc{Duration} timeout not exceeded \textbf{do} \\
         \ \,\,\,5: \> \>\quad\= $c_{\mathrm{best}}$ \quad\,\,\,\,\= $\gets$ \,\, $D$.\textsc{GlobalBestCost}() \\
         \ \,\,\,6: \> \>\quad\= $c_{\mathrm{cur}}$ \>$\gets$ \,\, $D$.\textsc{Cost}($s_{\mathrm{cur}}$) \\
         \ \,\,\,7: \> \> \>$c_{\mathrm{new}}$ \>$\gets$ \,\, $D$.\textsc{ApplyLLH}($H$.\textsc{LLH},\,$s_{\mathrm{cur}}$,\,$s_{\mathrm{new}}$) \\
         \ \,\,\,8: \> \> \> \textbf{if} $c_{\mathrm{new}}$ $<$ $c_{\mathrm{cur}}$ \textbf{then} \\  

         \ \,\,\,9: \> \> \> \quad \= \boldsymbol{$n_{\mathrm{imp}}$} \= $\gets$ \,\, \= \boldsymbol{$n_{\mathrm{imp}}$} + 1 \\
         \ \,\,10: \> \> \> \> \boldsymbol{$\mu$} \> $\gets$ \> $\boldsymbol{\mu}+(c_{\mathrm{cur}}-c_{\mathrm{new}}-\boldsymbol{\mu}) / \boldsymbol{n_{\mathrm{imp}}}$ \\
         \ \,\,11: \> \> \>\textbf{if} $H$.\textsc{Accept}(\boldsymbol{$\mu$}, $c_{\mathrm{best}}$, $c_{\mathrm{cur}}$, $c_{\mathrm{new}}$) \textbf{then} \\
         \ \,\,12: \> \> \> \quad $D$.\textsc{CopySolution}($s_{\mathrm{new}}$, $s_{\mathrm{cur}}$) \\
         \ \,\,13: \> \>$D$.\textsc{RestorePerturbationIntensity}() \\
         \ \,\,14: \> \>$D$.\textsc{CopySolution}($s_{\mathrm{cur}}$, $s^{**}$) \\
         \ \,\,15: \> \>\textbf{return} $D$.\textsc{Cost}($s^{**}$)
     \end{tabbing}
     \caption{Application of virtual LLH.}
     \label{code:pseudocode_apply_virtual_heuristic}
\end{figure}
Figure \ref{code:pseudocode_apply_virtual_heuristic} shows how the three modifiers take effect when applying a virtual LLH. On the highest level, the procedure \textsc{ApplyVirtualLLH} keeps an interface identical to the standard LLH application in HyFlex. The virtual LLH $H$ is applied on a solution stored in the source solution register $s^{*}\!$, the resulting solution is saved in the target solution register $s^{**}\!$, and the cost of this new solution is returned. The repeated application of the wrapped LLH linked to the \textsc{Duration} attribute is realized by means of the while loop on line 4. This aspect allows for executing the given original LLH for a certain amount of time repeatedly. Modification in terms of perturbation intensity related to the \textsc{Intensity} attribute is realized on lines 3 and 13. Line 3 instructs the domain to execute perturbative LLHs in the main loop with the intensity required by the virtual LLH (if applicable for the LLH type). At the end of the virtual LLH application, the original perturbation intensity is restored at line 13. 

The most involved modifier is the acceptance strategy provided in the \textsc{Accept} attribute. Its key role is to discard new solutions after the LLH application if their quality is deemed insufficient by the standards of the given strategy. This allows for keeping the search trajectory within promising parts of the search space. In order to accommodate this modifier, we introduce two additional solution registers, $s_{\mathrm{cur}}$ and $s_{\mathrm{new}}$. The register $s_{\mathrm{cur}}$ holding the current solution to search from is initialized on line 2 by a copy of the solution in $s^{*}\!$. Then, line 7 inside the repetition loop applies the original LLH to this solution and saves the new solution to $s_{\mathrm{new}}$. The acceptance strategy then decides whether the solution in $s_{\mathrm{new}}$ is of sufficient quality to be copied to $s_{cur}$ on lines 11 and 12, or whether the solution in $s_{\mathrm{new}}$ shall be discarded. Ultimately, the last accepted solution from $s_{\mathrm{cur}}$ is copied to the target solution register $s^{**}$ on line 14 as the result of the virtual LLH application, and its quality is returned on line 15. We note that, in general, the acceptance strategy decisions are based on the qualities of the globally best-so-far (line 5), current (line 6), and new (line 7) solutions. The last input to the acceptance decision is the global \textit{mean improvement} statistic denoted as \boldsymbol{$\mu$}. This statistic is critical for the search acceptance strategies to be able to operate under the cross-domain context. This global statistic is initially set to 0 and then continuously updated by means of lines 8 to 10. 

\subsection{LLH transformation principles}

Next, we detail and motivate the three key principles and propose novel cross-domain acceptance strategies utilizing the \boldsymbol{$\mu$} statistic.

\subsubsection{Solution acceptance}

Acceptance methods such as threshold acceptance~\cite{ACCEPTANCE_THRESHOLD} (TA), record-to-record travel~\cite{ACCEPTANCE_R2R} (R2R), or Metropolis acceptance~\cite{ACCEPTANCE_METROPOLIS} (MA) are common search-steering mechanisms in the problem-specific context. However, these methods face three critical issues in the context of cross-domain search. First, (1) the ranges and granularity of objective function values vastly differ across problem domains, and (2) across instances inside the individual domains. Moreover, (3) the objective function values within one search run may differ by several orders of magnitude. To the best of our knowledge, FSILS is the only work explicitly describing at least the points (1) and (2) as fundamental issues in cross-domain search. Importantly, FSILS proposes MA with a modification resolving not only the issues (1) and (2) described in their paper, but also the issue (3). The modification's key point is the normalization of the solution qualities using the mean improvement statistic $\mu$. We adopt this $\mu$-normalization technique with the difference that we update the statistic after every improvement, compared to the sparser updates at the end of LS chains in FSILS, resulting in more precise estimates. Critically, instead of just copying the FSILS design and $\mu$-normalized MA as is common (see Table~\ref{table:hh_properties}), we observe that $\mu$-normalization solves a general cross-domain issue and can therefore be used with other common acceptance methods, not only with MA. Consequently, we propose novel cross-domain acceptance mechanisms by $\mu$-normalizing standard problem-specific TA, R2R, and MA acceptances, all three with both constant (CONST) and exponential (EXP) threshold cooling schedules. 

Equations \ref{eq:metropolis} to \ref{eq:r2r} describe the acceptance criteria MA, TA, and R2R, respectively. $\tau$ is their threshold parameter. $\mathit{UNIFORM}(0, 1)$ denotes a random choice from a uniform distribution on the interval 0 to 1. We note that in all strategies, strictly improving solutions are always accepted.
\begin{align}
    \mathit{UNIFORM}(0, 1) & < e^{\frac{c_{\mathrm{cur}}-c_{\mathrm{new}}}{\tau \mu}} \label{eq:metropolis} \\
    c_{\mathrm{new}} & \leq c_{\mathrm{cur}} \,\, + \tau \mu  \label{eq:threshold} \\
    c_{\mathrm{new}} & \leq c_{\mathrm{best}} + \tau \mu \label{eq:r2r}
\end{align}
The CONST variants set $\tau$ to a fixed value. The EXP variants decrease $\tau$ over time based on two initial parameters, $\tau_{\mathrm{start}}$ and $\tau_{\mathrm{end}}$. We fit an exponential function $f$ of base $e$ such that $f(0) = \tau_{\mathrm{start}}$ and $f(1) = \tau_{\mathrm{end}}$. During the search, the effective $\tau$ is calculated as $f(x)$ where $x \in [0;1]$ is the proportion of already consumed search budget. We note that the CONST variant of MA matches the acceptance from FSILS.

\subsubsection{LLH repetitions}

We use the timeout-based repetitions introduced in LUBY for two reasons. First, with our proposed cross-domain acceptance mechanisms, the original (overly aggressive) discarding of worse solutions in domain amplifications can be replaced, allowing for the formerly advertised "collaboration among heuristics". Second, we observe that replacing the original LLHs such that \textit{all} new LLHs have the \textit{same} (high-enough) repetition timeout implicitly normalizes the computational resources allocated to individual LLHs (if no other bias is present). This contrasts with the original domain amplifications that double the LLH set by keeping the original LLHs. We note that projecting the speed of LLHs into their sampling probabilities has been shown as beneficial, e.g., in the \textit{SpeedNew} mechanism from FSILS.

\subsubsection{Perturbation intensity}

Based on the recent meta-study of the hyper-heuristics competing in CHeSC~\cite{METASTUDY_2025}, 8 of the 20 methods use a static setting of the perturbative LLH intensity parameter. Similarly, the post-CHeSC methods summarized in Table~\ref{table:hh_properties} also often do not work with this parameter. At the same time, one of the conclusions of the aforementioned meta-study is that the static parameter setting is the only parameter control scheme inferior to other alternatives that are rather interchangeable in terms of performance. Here, our observations suggest that the critical point is primarily the ability to vary the perturbation intensity. Therefore, we propose duplication of perturbative LLHs based on a predefined set of variable intensity settings. This transformation effectively offers perturbations of varying intensities with the possibility to bias the sampling probabilities of more/less aggressive perturbations. Moreover, such a transformation can be used as an arguably simpler replacement for existing parameter adaptation mechanisms.

\section{Experiments}

Now, we experimentally demonstrate the benefits of strategic LLH set transformations. The first part of the experiments is based on the standard CHeSC benchmark. For 3 hyper-heuristics, we gradually construct complete LLH set transformations by sequentially transforming solution acceptance, LLH repetitions, and perturbation intensities. Further, we improve 4 additional hyper-heuristics by transforming only the intensity of perturbations, effectively replacing their internal intensity adaptation mechanisms. In the second part of the experiments, we take the developed transformations and show their generality on three real-world domains.

\subsubsection{Methods}

The 3 hyper-heuristic strategies that we gradually built upon are the following. First, we concentrate on the recent MC hyper-heuristic utilizing Q-learning LLH selection, Luby sequence restarts, and domain amplifications. Second, we inspect the closely related hyper-heuristic LUBY using only the Luby sequence restarts, domain amplifications, and uniform random selection of LLHs. Lastly, we introduce a baseline naive hyper-heuristic (NHH) that first uniformly randomly selects an LLH category (LS, RR, or MUT) and then uniformly randomly selects an LLH within this category, while always accepting new solutions. For method $\textrm{X}\in\{\textrm{MC}, \textrm{LUBY}, \textrm{NHH}\}$, we use the notation X\textsuperscript{+} and X\textsuperscript{0} to explicitly distinguish variants of X with domain amplification and without it. This choice of methods allows us to (1) assess raw effects of the transformations while avoiding interactions with common design biases using NHH, (2) improve MC as a non-trivial learning-based method, and (3) inspect LUBY as its logical subset, dropping MC's learning component. Furthermore, MC and LUBY are the only existing methods using the original domain amplifications, allowing for their direct comparison with our LLH set transformations. The 4 hyper-heuristics where we transform only the perturbation intensities are LGIHH, FSILS, TSILS, and EAILS. Lastly, we note that we also compare with FRAMAB, LASTRL, QHH, MCTS, and GEPHH on the CHeSC benchmark based on the results reported by the authors. In case of NHH, MC, LUBY, TSILS, EAILS, FSILS, LGIHH, and all their derived variants, we provide our own reevaluations under the same conditions.

\subsection{CHeSC benchmark experiments}

We use the standard cross-domain CHeSC benchmark and the connected HyFlex framework \cite{HYFLEX_benchmark} implementing 6 diverse search domains. Namely, the domains are Maximum Satisfiability, Bin Packing, Personnel Scheduling, Flowshop, Travelling Salesman Problem, and Vehicle Routing Problem. Our experiments are based on the 30 instances used in the original competition (5 for each domain), allowing for extensive comparison with existing hyper-heuristics. To quantify the gradual improvements during the sequential construction of LLH set transformations, we compare the methods following the rules and F1 scoring system used in CHeSC. Specifically, we calculate the F1 score for a method by letting it compete against the original results reported for the 20 hyper-heuristics competing in CHeSC. Higher F1 scores reflect better results. We perform 31 repeated evaluations of each of the 30 competition instances. The competition benchmark script allocated 276 seconds to one run in our environment running on \textit{Debian 6.1.135 x86\_64} using \textit{AMD EPYC 7543} CPU ($2.8$ GHz), matching a 10-minute timeout on the original competition machine. The HyFlex code was compiled using \textit{openjdk 11.0.27}. Further details about the domains, LLH sets, instances, and F1 scoring are described in \cite{CHeSC_competition}. Detailed results of these experiments are in Appendix A.

\begin{table}[tb]
    \begin{center}
    \small
    \begin{tabular}{cccrrrr}
        & \multicolumn{1}{l}{\textit{Variant}} & \textit{Acceptance} & \multicolumn{1}{l}{$\tau_{\mathrm{min}}$} & \multicolumn{1}{l}{$\tau_{\mathrm{max}}$} & \multicolumn{1}{l}{$\tau_{\mathrm{step}}$} & $\tau_{\mathrm{end}}$ \\
        
        \multirow{6}{*}{\rotatebox[origin=l]{90}{\textbf{NHH}}} & \multirow{3}{*}{CONST}    & R2R & 1.0 & 6.0 & 1.25 & -- \\
                 & & MA & 0.25 & 1.25 & 0.25 & -- \\
                 & & TA & 0.25 & 1.25 & 0.25 & -- \\[0.2em]
        
        & \multirow{3}{*}{EXP}      & R2R & 2.5 & 7.5 & 1.25 & 1.0 \\
                                  & & MA & 0.5 & 1.5 & 0.25 & 0.25 \\
                                  & & TA & 0.5 & 1.5 & 0.25 & 0.25 \\[0.3em]

           \multirow{6}{*}{\rotatebox[origin=l]{90}{\textbf{MC+LUBY}}} & \multirow{3}{*}{CONST} & R2R & 1.0 & 6.0 & 1.25 & -- \\
                                  & & MA & 0.25 & 2.25 & 0.5 & -- \\
                                  & & TA & 1.0 & 2.0 & 0.25 & -- \\[0.2em]
            & \multirow{3}{*}{EXP}   & R2R & 2.5 & 12.5 & 2.5 & 1.0 \\
                                  & & MA & 0.75 & 2.75 & 0.5 & 0.25 \\
                                  & & TA & 1.25 & 2.25 & 0.25 & 1.0
    \end{tabular}
    \vspace*{-1.5mm}
    \caption{
        Summary of parameter scales (5 values) for each combination of hyper-heuristic, acceptance, and variant. \\
        CONST: $\tau$ from $\tau_{\mathrm{min}}$ to $\tau_{\mathrm{max}}$ with a step $\tau_{step}$. \\
        EXP: $\tau_{\mathrm{start}}$ from $\tau_{\mathrm{min}}$ to $\tau_{\mathrm{max}}$ with a step $\tau_{step}$. $\tau_{\mathrm{end}}$ is fixed.
    }
    \label{table:acceptance_params}
    \vspace*{-1.5mm}
    \end{center}
\end{table}

\subsubsection{Solution acceptance}

We test transforming the LLHs in both RR and MUT categories with one of the 6 cross-domain acceptance strategies, i.e., MA, TA, and R2R, all in both CONST and EXP variants. We test 5 increasingly strict parameterizations for each strategy. The respective parameter ranges were identified with preliminary experiments. Table~\ref{table:acceptance_params} summarizes the selected acceptance parameters used with NHH\textsuperscript{0}, LUBY\textsuperscript{0}, and MC\textsuperscript{0}. We note that the higher $\tau$ values in LUBY\textsuperscript{0}/MC\textsuperscript{0} reflect their regular intensifying restarts to the best-so-far solution missing in NHH. Now, we evaluate the outlined setups and summarize our main observations.

Crucially, all of the 6 acceptance strategies, given a reasonable setting, provide benefits in terms of the method's cross-domain performance. For all NHH, LUBY, and MC, the best acceptance transformations outperform the respective X\textsuperscript{0} and X\textsuperscript{+} setups. Even more, this is the case for a large part of the tested configurations across all 6 acceptance strategies for NHH and LUBY. These results provide strong evidence that (1) a properly set acceptance mechanism is a design element of critical importance for the overall cross-domain performance, and (2) $\mu$-normalization can be successfully used to derive novel cross-domain acceptance strategies as we propose. In relative comparison, we observe the most consistent performance in the R2R EXP strategy. It achieves the best results for MC and LUBY, and the second-best results, with 1 F1 point difference from TA EXP, for NHH. We provide two notable observations regarding R2R EXP. First, R2R strategies establish a global bound on what is (un)acceptable solution quality, contrasting with restrictions on a single acceptance step in MA and TA. Second, the EXP variant introduces an interesting advantage in the cross-domain context. While domains generally differ in their ideal setting of $\tau$, exponential decay of $\tau$ ensures that searching under different settings of $\tau$ takes place. Interestingly, the experiments also revealed one important weakness of the TA strategy in the cross-domain context. Generally, we observe that the deteriorating steps should be kept smaller than $\mu$, i.e., $\tau < 1.0$, at least for a larger part of the search process. At the same time, some domains (e.g., Maximum Satisfiability) have the inherent property of atomic improvement units (unsatisfied clauses). In case $\mu$ is close to 1.0, its combination with $\tau < 1.0$ often leads to premature inability to accept even small qualitative deterioration. Based on the aforementioned observations, we fix the acceptance transformations to R2R EXP with $\tau_{start}$ set to 5.0, 7.5, and 10.0 for NHH\textsuperscript{0}, LUBY\textsuperscript{0}, and MC\textsuperscript{0}, respectively. We refer to the resulting methods as NHH\textsuperscript{A}, LUBY\textsuperscript{A}, and MC\textsuperscript{A}.

\subsubsection{LLH repetitions}

For NHH\textsuperscript{A}, LUBY\textsuperscript{A}, and MC\textsuperscript{A}, we further introduce repeated LLH applications for the categories LS, RR, and MUT. We test 5 distinct repetition timeouts: 0.5, 1, 2.5, 5, and 10 milliseconds. Regarding the results, enforcing repetition timeouts on LLHs further boosted the performance of NHH\textsuperscript{A}, LUBY\textsuperscript{A}, and MC\textsuperscript{A} in all of the tested repetition timeouts. Generally, shorter timeouts resulted in better benefits. Longer timeouts converge back towards the results of NHH\textsuperscript{A}, LUBY\textsuperscript{A}, and MC\textsuperscript{A}. We also performed an ablation analysis separating the repetitions' effects on the LS category from effects on the perturbative RR and MUT categories. We conclude that both LS and RR+MUT parts separately add to the transformation's performance. However, transforming all LS, RR, and MUT provides better performance than each of the components separately. We fix the best repetition configurations to 0.5 ms for LUBY\textsuperscript{A} and MC\textsuperscript{A}, and 1 ms for NHH\textsuperscript{A}. We refer to the resulting methods as NHH\textsuperscript{AR}, LUBY\textsuperscript{AR}, and MC\textsuperscript{AR}. 

\subsubsection{Perturbation intensity}

Apart from extending NHH\textsuperscript{AR}, LUBY\textsuperscript{AR}, and MC\textsuperscript{AR}, we also modify (only) the perturbation intensities of LGIHH, TSILS, FSILS, and EAILS. While these methods handle acceptance and repetition aspects reasonably, they either do not handle intensities (FSILS) or use adaptive mechanisms that can be potentially overriden with substantially simpler alternatives (LGIHH, EAILS, TSILS). In the considered transformations, we target LLHs in the RR and MUT categories and duplicate their LLHs in several intensities. Setup $I$ = [0.1,\,0.2...0.9,\,1.0] uniformly covers the whole parameter scale. Setup \textit{II} = [0.05,\,0.05,\,0.05,\,0.05,\,0.1,\,0.1,\,0.2,\,0.3,\,0.5] covers the lower half of the parameter scale with an exponential ramp up, prioritizing low intensities. Setup \textit{III} = [0.1,\,0.2,\,0.3] allows for slight deviations from the HyFlex default 0.2. Setup \textit{Base} refers to X\textsuperscript{AR} (NHH, LUBY, MC) or X\textsuperscript{0} (LGIHH, FSILS, TSILS, EAILS). The results are summarized in Table~\ref{table:hyflex_intensities}.

\begin{table}[t]
    \begin{center}
    \small
    \begin{tabular}{lccccccc}
        \rotatebox[origin=l]{0}{} & \rotatebox[origin=l]{0}{\hspace*{-1.5mm}NHH} & \rotatebox[origin=l]{0}{\hspace*{-1.85mm}LUBY} & \rotatebox[origin=l]{0}{MC} & \rotatebox[origin=l]{0}{\hspace*{-1.85mm}LGIHH} & \rotatebox[origin=l]{0}{\hspace*{-1.85mm}FSILS} & \rotatebox[origin=l]{0}{\hspace*{-1.85mm}TSILS} & \rotatebox[origin=l]{0}{\hspace*{-1.85mm}EAILS} \\
        \hline \\[-0.75em]
    \textit{Base}   &         115  & 154 &         164  &         210  &        193  &         220  &         187  \\
    \textit{I}      &          94  &  82 &         130  & \textbf{213} & \textbf{203} &         203  &         142  \\
    \textit{II}     & \textbf{154} & 143 & \textbf{192} & \textbf{225} &         192  &         210  &         179  \\
    \textit{III}    & \textbf{116} & 145 & \textbf{191} & \textbf{214} & \textbf{202} & \textbf{224} & \textbf{193} \\[-0.5em]
    \end{tabular}
    \caption{
        F1 scores for setups \textit{I}-\textit{III} and \textit{Base} (X\textsuperscript{AR} or X\textsuperscript{0}) for the tested 7 hyperheuristics. \textbf{Bold}: outperforms \textit{Base}. 
    }
    \label{table:hyflex_intensities}
    \end{center}
\end{table}
Overall, the intensities clearly show as an important performance leverage. All methods except for LUBY\textsuperscript{AR} can be improved with at least one (but often multiple) setups \textit{I}-\textit{III}. Generally, we see the more conservative setups \textit{II} and \textit{III} as a better choice than \textit{I}. The most consistent is the setup \textit{III}. The setup \textit{II} works well with methods not using the ILS scheme (see Table~\ref{table:hh_properties}). Lastly, the more aggressive setup \textit{I} rather deteriorates performance, an exception is methods using full search restarts (FSILS, LGIHH). Based on the results, we fix the setup to \textit{II} for NHH\textsuperscript{AR}, MC\textsuperscript{AR}, and LGIHH, use the setup \textit{III} for LUBY\textsuperscript{AR}, TSILS, and EAILS, and fix the setup \textit{I} for FSILS. The resulting methods are referred to as X\textsuperscript{*}.

\subsubsection{Results summary}

\begin{table}[t]
    \begin{center}
    \small
    \begin{tabular}{lccccccc}
        \rotatebox[origin=l]{0}{} & \rotatebox[origin=l]{0}{\hspace*{-2mm}NHH} & \rotatebox[origin=l]{0}{\hspace*{-2mm}LUBY} & \rotatebox[origin=l]{0}{\hspace*{-1mm}MC} & \rotatebox[origin=l]{0}{\hspace*{-2mm}LGIHH} & \rotatebox[origin=l]{0}{\hspace*{-2mm}FSILS} & \rotatebox[origin=l]{0}{\hspace*{-2mm}TSILS} & \rotatebox[origin=l]{0}{\hspace*{-2mm}EAILS} \\
        \hline \\[-0.75em]
    X\textsuperscript{0}   &   1  &  69 & 105 & 210 & 193 & 220 & 187 \\
    X\textsuperscript{+}   &  14  &  82 & 137 &  -- &  -- &  -- &  -- \\
    X\textsuperscript{A}   &  94  & 136 & 149 &  -- &  -- &  -- &  -- \\
    X\textsuperscript{AR}  & 115  & 154 & 164 &  -- &  -- &  -- &  -- \\
    X\textsuperscript{*}   & 154  & 145 & 192 & 225 & 203 & 224 & 193
    \end{tabular}
    \caption{
        Summary of F1 scores for 7 hyper-heuristics with (gradually) constructed LLH set transformations.
    }
    \label{table:hyflex_summary}
    \vspace*{-2.5ex}
    \end{center}
\end{table}

Table~\ref{table:hyflex_summary} summarizes the F1 scores obtained for the incrementally constructed LLH set transformations in all of the 7 tested methods. The proposed transformations successfully boosted F1 scores for all of the 7 methods. For LGIHH, FSILS, TSILS, and EAILS, overriding (or adding the missing) intensity handling via transformed LLH sets systematically provides benefits. For LGIHH and TSILS, the results even outperform the original TSILS performance, setting the new state-of-the-art on the CHeSC benchmark. This overall suggests that manipulating the LLH sets offers a simpler and more transparent alternative to the parameter adaptation mechanisms present in the original methods. Regarding NHH, LUBY, and MC, we can see substantial benefits of the best transformations compared to both the X\textsuperscript{0} and X\textsuperscript{+} baseline variants. In the original CHeSC competition, NHH\textsuperscript{*} and LUBY\textsuperscript{*} would rank 2nd with results close to the winning GIHH, and MC\textsuperscript{*} would score 1st. Strikingly, \textit{NHH\textsuperscript{*} is a trivial unbiased LLH selection mechanism, only set to operate in a reasonably safe environment (acceptance) with normalized granularity at which new LLHs are sampled (repetitions), allowing the perturbation intensity to vary}. Crucially, this result implies that the key properties of the LLH set play a comparable, if not more important, role than the selection mechanism. In this regard, we contribute a general tool for controlling LLH aspects with critical impacts on the overall performance.

\begin{landscape}

\begin{table*}[t]
\centering
\small
\begin{tabular}{l @{\hspace*{0.5mm}} c @{\hspace*{3.0mm}} c @{\hspace*{3.0mm}} c @{\hspace*{3.0mm}} c @{\hspace*{3.0mm}} c @{\hspace*{3.0mm}} c @{\hspace*{3.0mm}} c @{\hspace*{3.0mm}} c @{\hspace*{3.0mm}} c @{\hspace*{3.0mm}} c @{\hspace*{3.0mm}} c @{\hspace*{3.0mm}} c}
X vs. & \textsc{NHH\textsuperscript{+}} & \textsc{LUBY\textsuperscript{+}} & \textsc{MC\textsuperscript{+}} & \textsc{LGIHH} & \textsc{FSILS} & \textsc{TSILS} & \textsc{EAILS} & \textsc{FRAMAB} & \textsc{LASTRL} & \textsc{QHH} & \textsc{MCTS} & \textsc{GEPHH} \\
\toprule
NHH\textsuperscript{*}   & \textbf{0.000} &     0.226 & 0.379 & 0.931 & 0.824 & 0.878 & 0.908 & \textbf{0.037} & 0.525 & 0.185 & 0.948 & 0.941 \\
LUBY\textsuperscript{*} & \textbf{0.000} & \textbf{0.004} & 0.059 & 0.899 & 0.792 & 0.982 & 0.605 & \textbf{0.004} & 0.617 & 0.103 & 0.906 & 0.725 \\
MC\textsuperscript{*}   & \textbf{0.000} & \textbf{0.000} & \textbf{0.003} & 0.686 & 0.625 & 0.878 & 0.311 & \textbf{0.002} & 0.245 & 0.067 & 0.533 & 0.410 \\
LGIHH\textsuperscript{*} & \textbf{0.000} & \textbf{0.000} & \textbf{0.002} & \textbf{0.037} & 0.075 & 0.275 & \textbf{0.003} & \textbf{0.000} & \textbf{0.004} & \textbf{0.000} & 0.109 & 0.289 \\
FSILS\textsuperscript{*} & \textbf{0.000} & \textbf{0.006} & \textbf{0.017} & 0.686 & 0.325 & 0.855 & 0.275 & \textbf{0.000} & 0.116 & \textbf{0.001} & 0.525 & 0.492 \\
TSILS\textsuperscript{*} & \textbf{0.000} & \textbf{0.000} & \textbf{0.002} & 0.164 & \textbf{0.015} & \textbf{0.020} & \textbf{0.000} & \textbf{0.000} & \textbf{0.002} & \textbf{0.000} & 0.064 & 0.272 \\
EAILS\textsuperscript{*} & \textbf{0.000} & \textbf{0.000} & \textbf{0.031} & 0.707 & 0.758 & 0.953 & 0.226 & \textbf{0.000} & 0.059 & 0.050 & 0.500 & 0.633 \\
\bottomrule
\end{tabular}
\vspace*{-1.25ex}
\caption{
    Overview of p-values ($<0.05$ in \textbf{bold}). Alternative hypothesis: "row key is better than column key".
}
\label{table:wilcoxon_hyflex}
\vspace*{-0.75ex}
\end{table*}
\begin{table*}
\centering
\small
\begin{tabular}{lr| lr | lr | lr | lr | lr | lr}
\toprule
\textbf{NHH\textsuperscript{*}} & \textbf{446} & FSILS & 463 & FSILS & 446 & FSILS & 451 & FSILS & 466 & \textbf{TSILS\textsuperscript{*}} & \textbf{449} & FSILS & 446 \\
FSILS & 445 & EAILS & 417 & EAILS & 412 & \textbf{LGIHH\textsuperscript{*}} & \textbf{422} & EAILS & 425 & FSILS & 441 & \textbf{EAILS\textsuperscript{*}} & \textbf{431} \\
EAILS & 395 & TSILS & 411 & TSILS & 405 & EAILS & 401 & TSILS & 416 & EAILS & 397 & EAILS & 393 \\
TSILS & 388 & LGIHH & 358 & \textbf{MC\textsuperscript{*}} & \textbf{391} & TSILS & 387 & LGIHH & 375 & TSILS & 386 & TSILS & 391 \\
LGIHH & 353 & \textbf{LUBY\textsuperscript{*}} & \textbf{356} & LGIHH & 347 & LGIHH & 349 & \textbf{FSILS\textsuperscript{*}} & \textbf{278} & LGIHH & 344 & LGIHH & 353 \\
MC\textsuperscript{+} & 244 & MC\textsuperscript{+} & 249 & MC\textsuperscript{+} & 254 & MC\textsuperscript{+} & 250 & MC\textsuperscript{+} & 263 & MC\textsuperscript{+} & 249 & MC\textsuperscript{+} & 247 \\
LUBY\textsuperscript{+} & 218 & LUBY\textsuperscript{+} & 219 & LUBY\textsuperscript{+} & 229 & LUBY\textsuperscript{+} & 227 & LUBY\textsuperscript{+} & 236 & LUBY\textsuperscript{+} & 222 & LUBY\textsuperscript{+} & 225 \\
LUBY & 153 & LUBY & 165 & LUBY & 157 & LUBY & 157 & LUBY & 173 & LUBY & 154 & LUBY & 153 \\
MC & 127 & MC & 131 & MC & 128 & MC & 124 & MC & 139 & MC & 127 & MC & 129 \\

\bottomrule
\end{tabular}
\vspace*{-0.75ex}
\caption{
    F1 scores on the real-world domains. In each column, one X\textsuperscript{*} method (in \textbf{bold}) competes against referential results.
}
\label{table:f1_real_world}
\vspace*{-2ex}
\end{table*}
\end{landscape}

Second, we provide statistical tests for our results following the approach taken in \cite{2018_QHH}. For each instance, the methods' median results are min-max normalized. We use the min\,\&\,max medians based on the original CHeSC results. Then, the normalized medians on the 30 competition instances are used to perform the Wilcoxon signed rank test for a pair of competing hyper-heuristics. Table~\ref{table:wilcoxon_hyflex} summarizes comparisons of the methods with the final LLH set transformations compared to the baselines and \textit{all} recent hyper-heuristics. First, we conclude that the benefits of NHH\textsuperscript{*}, LUBY\textsuperscript{*}, and MC\textsuperscript{*} compared to their X\textsuperscript{+} counterparts are in all cases statistically significant (all X\textsuperscript{0} are worse than X\textsuperscript{+}). Among the remaining methods, LGIHH\textsuperscript{*} and TSILS\textsuperscript{*} benefits are statistically significant. For FSILS and EAILS, the p-values are inconclusive. Interestingly, the improvements from MC\textsuperscript{+} to MC\textsuperscript{*} make it competitive with the majority of recent methods.

\subsection{Real-world domains experiments}

The second part of the experiments applies the developed LLH set transformations to three additional real-world domains, validating their generality. The first domain is a rich variant of the pickup-delivery problem with time windows (PDPTW) arising from a freight-transportation application~\cite{alns_our_solver}. The objective is to minimize the travel distance and driver overtimes. The testing dataset consists of 18 instances, each with around 100 pickup-delivery transportation requests based on customer orders realized in the company Wereldo. We evaluate each instance 31 times with a 5-minute timeout per run, reflecting the typical use case for the solver. The other two domains have been described by \citet{kletzander2024hyper}. The second domain deals with the minimum shift design (MSD) problem. MSD aims to design shifts according to a given set of shift types such that a given demand for up to 50 employees working at each time slot with a granularity as low as 15 minutes is covered for a whole week. The objective is to cover the demand while minimizing the number of different pairs of shift starts and ends and the deviation from a target average shift length. We evaluate the same set of 33 realistic instances, each 5 times with a 60-minute timeout per run. The third domain is bus driver scheduling (BDS). In BDS, drivers are assigned to predetermined bus tours according to a complex set of constraints regarding limits of assignments and required breaks. A linear combination of several objectives is optimized. We evaluate 20 realistic instances with up to 1,000 bus legs. Each instance is evaluated 5 times with a 60-minute timeout per run. PDPTW runs are evaluated on \textit{Debian 6.1.135 x86\_64} using \textit{AMD EPYC 7543} CPU ($2.8$ GHz). MSD and BDS runs are evaluated on \textit{Ubuntu 22.04.2 LTS} with \textit{Intel Xeon E5-2650 v4} processors ($2.2$ GHz). We provide further details about the individual domains and their implementation details (available LLHs, perturbation intensity handling) in Appendix B.

Regarding comparisons, we again use the F1 scoring system and Wilcoxon tests. We evaluate and use LUBY, LUBY\textsuperscript{+}, MC, MC\textsuperscript{+}, LGIHH, TSILS, EAILS, and FSILS as the referential results. We omit NHH and NHH\textsuperscript{+} as both generally struggle to obtain feasible solutions. The referential results are used as the competitors in the F1 scoring and for both min-max normalization and comparisons with the Wilcoxon tests. Detailed results are in Appendix C.

\subsubsection{Evaluation} 

When switching from the CHeSC to the real-world domains, we encountered only one important difference related to the repeated LLH applications. Compared to the CHeSC domains, the real-world problems are more complex in terms of constraints and their evaluation. As a result, the LLHs have longer execution times, prolonging the repetition timeouts needed for the desirable effects. This shift is systematic and roughly one order of magnitude. Thus, we reflect the shift by a rough adjustment of the repetitions factor in NHH\textsuperscript{*}, LUBY\textsuperscript{*}, and MC\textsuperscript{*} to 10 ms. With this only change, we evaluate our X\textsuperscript{*} methods. Table~\ref{table:f1_real_world} summarizes the F1 scores of X\textsuperscript{*} competing against the referential results.

First, all methods X\textsuperscript{*} using the proposed LLH set transformations clearly outperform their X\textsuperscript{0} and X\textsuperscript{+} counterparts with one exception. For FSILS, the formerly inconclusive effects of intensity modification \textit{I} turned out to degrade the method's performance. The results thus tightly copy the key trends observed for the benchmark domains. Regarding the statistical significance, all observed improvements are significant, with the exception of EAILS\textsuperscript{*} having the p-value of 0.051. Second, we perform an ablation analysis for NHH\textsuperscript{*}, LUBY\textsuperscript{*}, and MC\textsuperscript{*} by sequentially adding the acceptance, repetition, and intensity transformations as for the CHeSC domains. We again confirm that each individual step adds to the overall performance of the transformations again replicating our observations from the CHeSC domains. Ultimately, we specifically emphasize the performance of NHH\textsuperscript{*}. While the results suggest that the hyper-heuristics dominating the CHeCS benchmark (FSILS, EAILS, TSILS, LGIHH) generalize well to the new domains, NHH\textsuperscript{*} outperforms all of these state-of-the-art methods. The only cases where NHH\textsuperscript{*} does not outperform its competitor with a statistically significant difference are FSILS and TSILS with Wilcoxon test p-values of 0.232 and 0.070. When comparing NHH\textsuperscript{*} with other X\textsuperscript{*} methods, we report p-values 0.466 for TSILS\textsuperscript{*}, 0.262 for EAILS\textsuperscript{*}, 0.119 for LGIHH\textsuperscript{*}, and p-values below the 0.05 threshold for FSILS\textsuperscript{*}, LUBY\textsuperscript{*}, and MC\textsuperscript{*}. In a closing remark, NHH\textsuperscript{*} also found several new best-known solutions, namely 3 for PDPTW and 8 for MSD, showing that simplicity and generality do not necessarily come at the expense of result quality.

\section{Conclusion}

Our paper identifies three critical principles affecting cross-domain search performance and exploits them to improve a wide range of existing hyper-heuristics. We demonstrate that the strategic transformations of LHH sets allow for outperforming the current state-of-the-art hyper-heuristics on both the standard CHeSC benchmark and three real-world domains. Strikingly, we achieve excellent results with a trivial random unbiased selection mechanism combined with a properly constructed set of LLHs. In this respect, we emphasize the conclusions of one of the pioneering works in hyper-heuristics~\cite{FISHER_THOMPSON_OLD}: "(1) an unbiased random combination of scheduling rules is better than any of them taken separately; (2) learning is possible". To (1), we add that a trivial unbiased random combination of LLHs may perform surprisingly well given the right set of LLHs. Regarding (2), we agree that learning \textit{how to select LLHs} is possible. Yet, we demonstrate and emphasize that the control over \textit{what LLHs we select} is comparably, if not more, important.

\newpage

\section*{Acknowledgments}

Computational resources were provided by the e-INFRA CZ project (ID:90254), supported by the Ministry of Education, Youth and Sports of the Czech Republic.

\noindent The research was supported by a grant from the OeAD-GmbH and AKTION as part of the scholarship programme Aktion Austria-Czech Republic.

\noindent  We would like to thank our industrial partner, Wereldo, for providing us with the real-world problem instances.

\noindent The financial support by the Austrian Federal Ministry for Digital and Economic Affairs, the National Foundation for Research, Technology and Development and the Christian Doppler Research Association is gratefully acknowledged.

\bibliography{bibliography}

\newpage

\end{document}